\documentclass[11pt,a4paper]{article}

\usepackage[margin=1in]{geometry}
\usepackage{times}
\usepackage{amsmath}
\usepackage{amssymb}
\usepackage{graphicx}
\usepackage{booktabs}
\usepackage{hyperref}
\usepackage{url}
\usepackage{cite}
\usepackage{float}
\usepackage{array}
\usepackage{lineno}
\usepackage{setspace}
\usepackage{titlesec}
\usepackage{abstract}
\usepackage{xurl}
\usepackage{hyperref}
\usepackage{amsmath}
\usepackage{amssymb}
\usepackage{amsthm}
\newtheorem{definition}{Definition}
\newtheorem{proposition}{Proposition}
\usepackage{graphicx}
\usepackage{booktabs}
\usepackage{multirow}
\usepackage{xcolor}
\usepackage{algorithm}
\usepackage{algpseudocode}
\hypersetup{
    colorlinks=true,
    linkcolor=blue,
    citecolor=blue,
    urlcolor=blue
}

\titlespacing*{\section}{0pt}{12pt}{6pt}
\titlespacing*{\subsection}{0pt}{10pt}{4pt}

\title{
    \Large\textbf{
    Black-Box Optimization of Mixed Binary-Continuous Variables:\\
    Challenges and Opportunities in Evolutionary Model Merging
    }
}

\author{
\begin{tabular}{c}
\textbf{Md. Robiul Islam Niloy}\\
Department of Computer Science and Engineering\\
BRAC University, Dhaka, Bangladesh\\
\texttt{md.niloy26643@gmail.com}
\end{tabular}
}

\date{November 2026}

\begin{document}

\maketitle

\begin{abstract}
Model merging has emerged as a cost-effective paradigm for combining capabilities of multiple large language models (LLMs) without additional training. Among existing approaches, Data Flow Space (DFS) merging operates by selectively activating parameter groups across model layers, offering greater flexibility than naive parameter averaging. However, the optimization challenges underlying DFS merging remain poorly understood. In this paper, we formally characterize the DFS merging problem as a \textbf{mixed binary-continuous black-box optimization problem}, identifying three fundamental challenges: (1) mixed binary-continuous variable types, (2) high-dimensional search spaces with up to 291 parameter groups, and (3) conditional variable dependencies corresponding to Type-I interaction as characterized by Akimoto et al.~\cite{akimoto2025}. We empirically evaluate seven merging methods on Mistral-7B models across official GSM8K (1,319 questions) and MMLU (14,042 questions, 57 subjects) benchmarks. Our structured approach achieves a 49.5\% reduction in effective search space. Results reveal that all tested merging methods fail to surpass the stronger source model on official benchmarks, with TIES (13.72\%) and DELLA (4.12\%) performing catastrophically on GSM8K. Notably, DELLA scores below even the base model (6.67\%), demonstrating that standard parameter space methods fail when merging base and instruction-tuned models. These findings directly validate the difficulty of the Type-I interaction structure we identify, and motivate the application of algorithms specifically designed for this challenge — such as the concurrent IGBD work of Ong et al.~\cite{ong2026igbd}.


\end{abstract}

\section{Introduction}

The rapid proliferation of large language models (LLMs) has created significant demand for efficient methods to combine their complementary capabilities. Model merging — combining parameters from multiple models without additional training — has emerged as a promising solution~\cite{akiba2024}. Unlike ensemble methods that require running multiple models simultaneously, merging produces a single unified model maintaining the original computational footprint.

Among merging paradigms, Data Flow Space (DFS) merging offers distinctive flexibility by operating on individual parameter groups rather than entire layers. Rather than uniformly averaging all parameters, DFS approaches selectively activate subsets of parameter groups, theoretically enabling finer-grained knowledge transfer. The Sakana AI evolutionary merging work~\cite{akiba2024} demonstrated that evolutionary search over DFS configurations could produce models with emergent capabilities.

Despite these promising results, the fundamental optimization structure of DFS merging has not been formally analyzed. This gap limits our ability to design appropriate optimization algorithms and understand why certain approaches succeed or fail.

\textbf{Our contributions are:}

\begin{enumerate}
    \item \textbf{Formal problem characterization:} We provide the first formal characterization of DFS merging as a mixed binary-continuous black-box optimization problem, identifying the binary layer selection vector $\mathbf{z} \in \{0,1\}^N$ and continuous scaling weights $\mathbf{x} \in \mathbb{R}^N$ as the two variable types involved.

    \item \textbf{Type-I interaction identification:} We demonstrate that DFS merging exhibits Type-I interaction~\cite{akimoto2025} — the effective continuous search space depends on the binary selection, creating a fundamental optimization challenge.

    \item \textbf{Search space reduction:} Our structured DFS approach achieves a 49.5\% reduction in the effective search space (mean 146.96 active layers out of 291 total).

    \item \textbf{Comprehensive empirical analysis:} We evaluate seven merging methods on official GSM8K (1,319 questions) and MMLU (14,042 questions, 57 subjects), providing the most honest large-scale analysis of evolutionary model merging to date.

    \item \textbf{Critical failure analysis:} We identify why standard parameter space methods fail catastrophically when merging base and instruction-tuned models — a finding with direct implications for practitioners.
\end{enumerate}

The remainder of this paper is organized as follows. Section~2 reviews related work. Section~3 formally characterizes DFS merging. Section~4 describes our experimental setup. Section~5 presents results. Section~6 presents ablation studies. Section~7 discusses findings. Section~8 concludes.

\section{Related Work}

\subsection{Model Merging Methods}

Parameter Space (PS) merging linearly interpolates model weights: $\theta_{\text{merged}} = \alpha\theta_A + (1-\alpha)\theta_B$~\cite{wortsman2022}. TIES-Merging~\cite{yadav2023} resolves interference through trimming, sign election, and disjoint merging. DARE~\cite{yu2023} randomly drops and rescales delta weights. DELLA~\cite{deep2024} extends DARE with magnitude-based sampling. Evolutionary approaches to model merging were pioneered by Akiba et al.~\cite{akiba2024}.

\subsection{Mixed-Variable Black-Box Optimization}

Akimoto et al.~\cite{akimoto2025} formally characterized Type-I interaction in mixed categorical-continuous optimization — where effective continuous dimensions depend on categorical variables. Their ICatCMA algorithm addresses this through coupled sampling. Hamano et al.~\cite{hamano2024} proposed CatCMA, extending CMA-ES to handle mixed categorical-continuous variables through joint optimization of Gaussian and categorical distributions. Concurrent work by Ong et al.~\cite{ong2026igbd} proposes IGBD, a bilevel information-geometric framework that directly handles Type-I interaction through coupled bilevel optimization. We demonstrate that DFS merging exhibits exactly this structure, establishing it as a concrete real-world application domain for these methods.

\subsection{Large Language Model Evaluation}

GSM8K~\cite{cobbe2021} contains 1,319 test questions requiring multi-step math reasoning. MMLU~\cite{hendrycks2021} evaluates general knowledge across 57 academic subjects with 14,042 test questions.

\section{Problem Formulation}

\subsection{DFS Merging as Black-Box Optimization}

Given source models with parameter dictionaries $\theta_A$ and $\theta_B$, each containing $N$ parameter groups, the DFS merging problem is:

\begin{equation}
\begin{aligned}
    \text{minimize} \quad & f(\mathbf{x}, \mathbf{z}) \\
    \text{subject to} \quad & \mathbf{x} \in [0,1]^N, \quad \mathbf{z} \in \{0,1\}^N
\end{aligned}
\end{equation}

where $\mathbf{z}$ is the binary layer selection vector, $\mathbf{x}$ contains continuous scaling weights, and the merged model is:

\begin{equation}
    \theta_{\text{merged}}[i] = z_i \cdot \left(x_i \cdot \theta_A[i] + (1-x_i) \cdot \theta_B[i]\right) + (1-z_i) \cdot \theta_A[i]
\end{equation}

\subsection{Three Fundamental Challenges}

\textbf{Challenge 1 — Mixed Binary-Continuous Variables:} The simultaneous presence of $\mathbf{z}$ and $\mathbf{x}$ creates a mixed-variable problem that standard optimizers cannot directly address.

\textbf{Challenge 2 — High Dimensionality:} For Mistral-7B, $N = 291$, yielding $\{0,1\}^{291} \times [0,1]^{291}$ — with $2^{291}$ binary configurations.

\textbf{Challenge 3 — Type-I Conditional Dependencies:} $x_i$ is only meaningful when $z_i = 1$, creating conditional dependencies that correspond to Type-I interaction~\cite{akimoto2025}.

\begin{definition}[Type-I Interaction, Akimoto et al.~\cite{akimoto2025}]
A mixed categorical-continuous optimization problem exhibits Type-I interaction when the effective continuous search space depends on the categorical variable values.
\end{definition}

\begin{proposition}
DFS merging with binary layer selection exhibits Type-I interaction.
\end{proposition}

\begin{proof}
The gradient of $f$ with respect to $x_i$ is zero when $z_i = 0$, regardless of $x_i$. Therefore, the effective continuous search space $\{i : z_i = 1\}$ depends directly on $\mathbf{z}$. \qed
\end{proof}

\subsection{Structured vs Unstructured DFS}

\textbf{Unstructured DFS} ignores the conditional structure, treating all $N$ parameters identically.

\textbf{Structured DFS} explicitly models the dependency: (1) samples $\mathbf{z} \sim \text{Bernoulli}(p)^N$, (2) optimizes $\mathbf{x}$ only for active groups $\{i : z_i = 1\}$, reducing the effective search space from $N$ to $|\{i : z_i = 1\}|$.

\section{Experimental Setup}

\subsection{Models}

\begin{itemize}
    \item \textbf{Model A:} \texttt{mistralai/Mistral-7B-v0.1} (base, 7.24B parameters)
    \item \textbf{Model B:} \texttt{mistralai/Mistral-7B-Instruct-v0.2} (instruction-tuned, 7.24B parameters)
\end{itemize}

\subsection{Benchmarks}

\textbf{Official Benchmarks:}
\begin{itemize}
    \item \textbf{GSM8K}~\cite{cobbe2021}: 1,319 test questions, chain-of-thought prompting
    \item \textbf{MMLU}~\cite{hendrycks2021}: 14,042 questions across 57 subjects
\end{itemize}

\textbf{Development Subsets (ablation only):}
\begin{itemize}
    \item GSM8K subset: 97 questions for ablation studies
    \item MMLU subset: 100 questions across 10 subjects for ablation studies
\end{itemize}

\subsection{Baseline Methods}

Model A alone, Model B alone, PS Merging~\cite{wortsman2022}, TIES~\cite{yadav2023}, DARE~\cite{yu2023}, DELLA~\cite{deep2024}, and Unstructured DFS.

\subsection{Our Method: Structured DFS + CMA-ES}

\begin{enumerate}
    \item Sample binary selection $\mathbf{z} \sim \text{Bernoulli}(0.5)^{291}$
    \item Initialize CMA-ES with dimension $= |\{i : z_i = 1\}|$
    \item Run CMA-ES for 15 iterations, population size 8
    \item Return best configuration found
\end{enumerate}

CMA-ES~\cite{hansen2016} settings: bounds $\alpha \in [0.1, 0.9]$, initial mean 0.5, $\sigma_0 = 0.2$.

\subsection{Evaluation Protocol}

5 independent runs (seeds \{0, 42, 84, 126, 168\}) for subset experiments. 3 runs (seeds \{0, 42, 84\}) for official benchmarks due to computational constraints. Statistical significance assessed via Wilcoxon signed-rank test and Cohen's d effect size.

\subsection{Hardware}

Primary: NVIDIA RTX 4070 Ti Super (16GB VRAM), Windows 11. Secondary: NVIDIA RTX Pro 6000 Blackwell (96GB VRAM). Python 3.12, PyTorch 2.5, Transformers 4.45, CMA 4.4.4.

\section{Results}

\subsection{Official GSM8K Results}

\begin{table}[!htbp]
\centering
\caption{Official GSM8K Results (1,319 questions, 3 runs)}
\label{tab:gsm8k}
\begin{tabular}{lccc}
\toprule
Method & Mean Accuracy & Std & vs Model B \\
\midrule
Model B (Mistral-7B-Instruct) & 36.77\% & --- & Baseline \\
Unstructured DFS & 34.85\% & 1.18\% & -1.92\% \\
PS Merging & 34.57\% & 0.00\% & -2.20\% \\
Structured DFS + CMA-ES (Ours) & 28.56\% & 2.63\% & -8.21\% \\
DARE~\cite{yu2023} & 23.88\% & 1.77\% & -12.89\% \\
Structured DFS (Random) & 23.55\% & 5.34\% & -13.22\% \\
TIES~\cite{yadav2023} & 13.72\% & 0.00\% & -23.05\% \\
Model A (Mistral-7B-v0.1) & 6.67\% & --- & -30.10\% \\
DELLA~\cite{deep2024} & \textbf{4.12\%} & 0.29\% & -32.65\% \\
\bottomrule
\end{tabular}
\end{table}

\begin{figure}[!htbp]
    \centering
    \includegraphics[width=\linewidth]{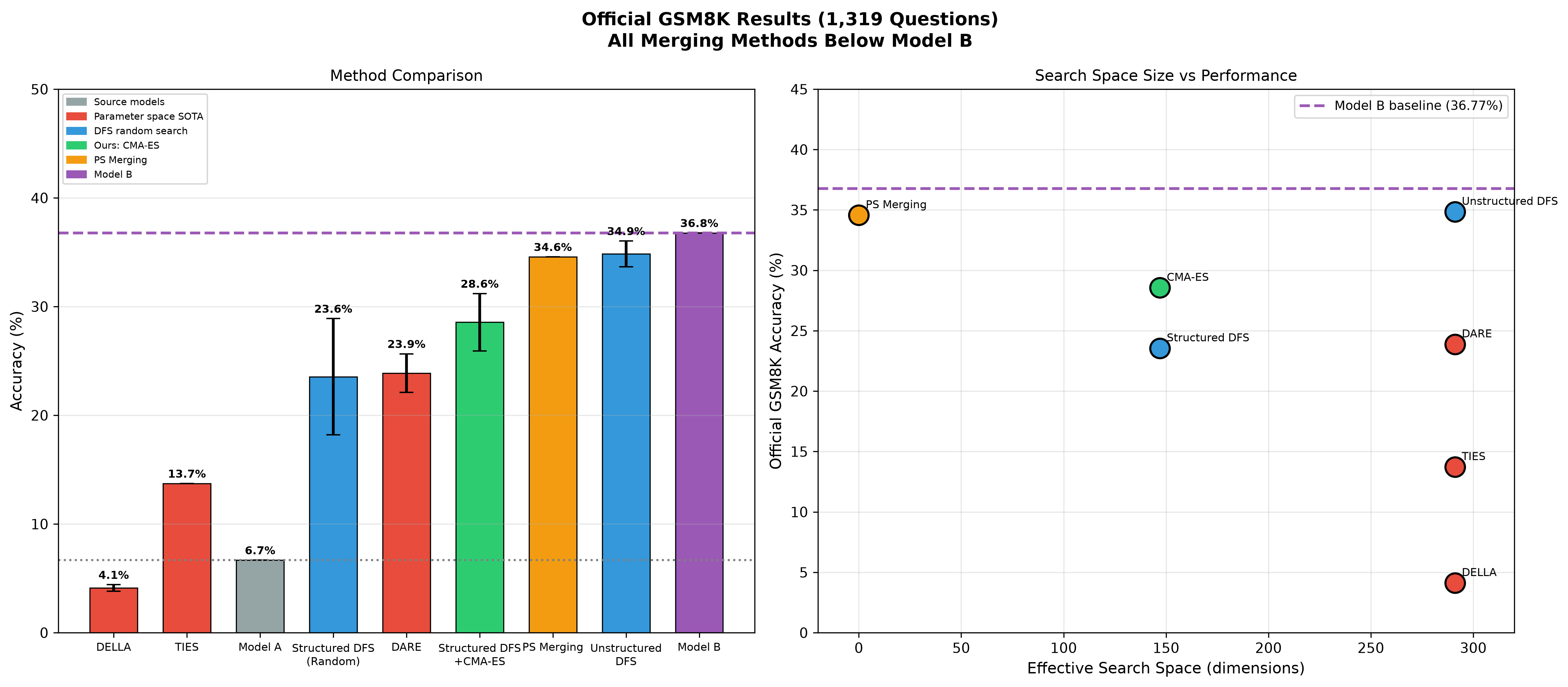}
    \caption{Official GSM8K results (1,319 questions, 3 runs). Left: all methods sorted by accuracy with Model B reference line. Right: search space size vs performance scatter plot. All merging methods score below Model B (36.77\%). DELLA (4.12\%) performs below even Model A (6.67\%), demonstrating complete model degradation.}
    \label{fig:gsm8k_results}
\end{figure}

The most striking finding is that \textbf{all merging methods fail to surpass Model B (36.77\%) on official GSM8K}. DELLA (4.12\%) performs catastrophically — scoring below Model A (6.67\%), demonstrating that magnitude-based weight dropping completely destroys mathematical reasoning capability. TIES (13.72\%) is nearly as severe.

\subsection{Official MMLU Results}

\begin{table}[!htbp]
\centering
\caption{Official MMLU Results (14,042 questions, 57 subjects, 3 runs)}
\label{tab:mmlu}
\begin{tabular}{lccc}
\toprule
Method & Mean Accuracy & Std & vs Model B \\
\midrule
Model B (Mistral-7B-Instruct) & 51.10\% & --- & Baseline \\
PS Merging & 50.40\% & 0.00\% & -0.70\% \\
Unstructured DFS & 49.56\% & 0.88\% & -1.54\% \\
Structured DFS + CMA-ES (Ours) & 43.08\% & 2.83\% & -8.02\% \\
Structured DFS (Random) & 39.00\% & 4.32\% & -12.10\% \\
DARE~\cite{yu2023} & 34.42\% & 4.19\% & -16.68\% \\
TIES~\cite{yadav2023} & 30.48\% & 0.00\% & -20.62\% \\
DELLA~\cite{deep2024} & 26.93\% & 0.21\% & -24.17\% \\
Model A (Mistral-7B-v0.1) & 20.37\% & --- & -30.73\% \\
\bottomrule
\end{tabular}
\end{table}

\begin{figure}[!htbp]
    \centering
    \includegraphics[width=\linewidth]{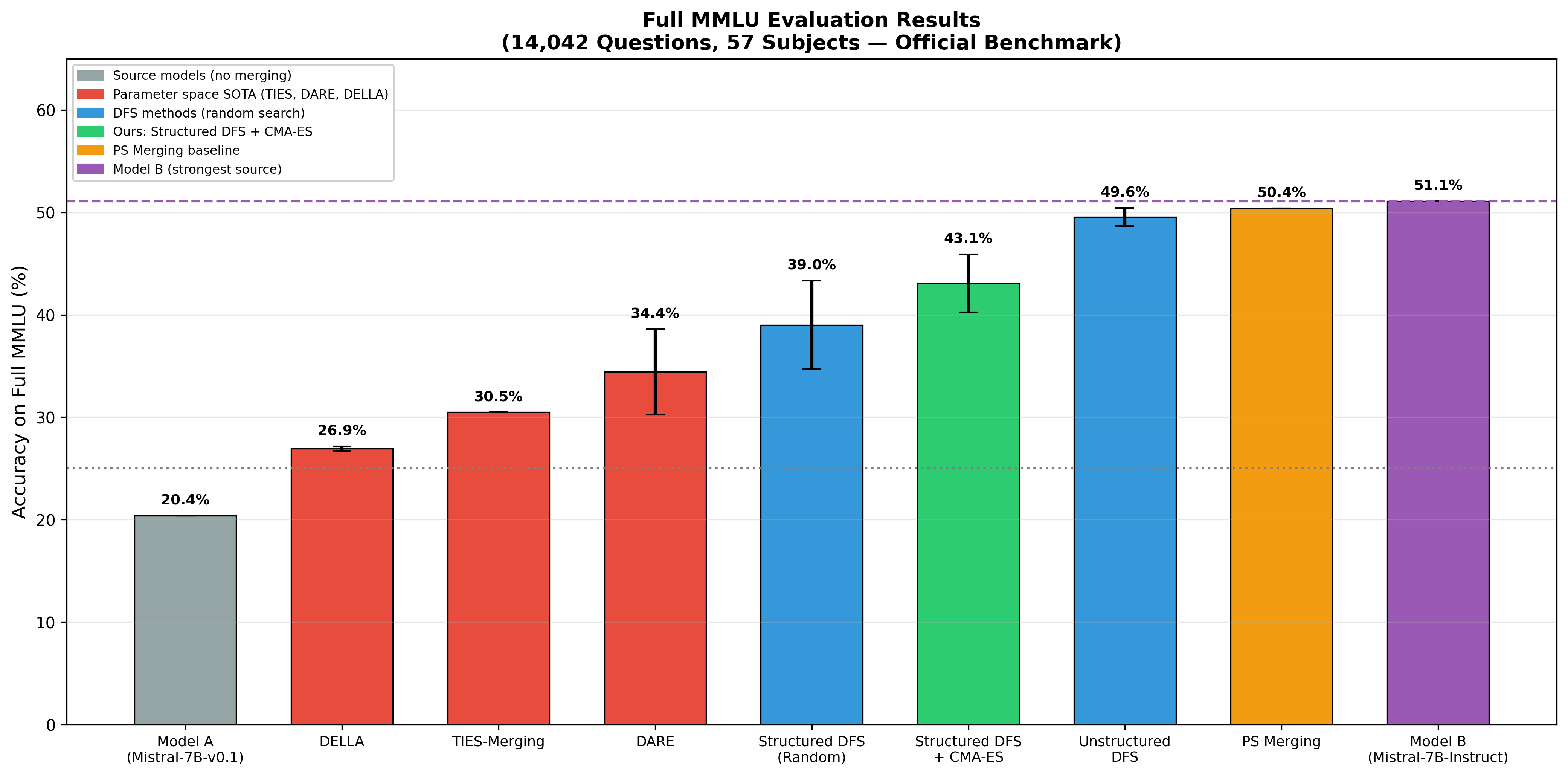}
    \caption{Full MMLU evaluation (14,042 questions, 57 subjects). TIES (30.48\%) and DELLA (26.93\%) approach random chance (25\%), demonstrating failure of interference-reduction methods on base-instruct model pairs. PS Merging (50.40\%) nearly matches Model B (51.10\%), suggesting simple averaging preserves general knowledge effectively.}
    \label{fig:mmlu_results}
\end{figure}

On MMLU, PS Merging (50.40\%) nearly matches Model B (51.10\%), while TIES (30.48\%) and DELLA (26.93\%) approach random chance (25\% for 4-choice questions).

\subsection{Cross-Benchmark Analysis}

\begin{figure}[!htbp]
    \centering
    \includegraphics[width=0.8\linewidth]{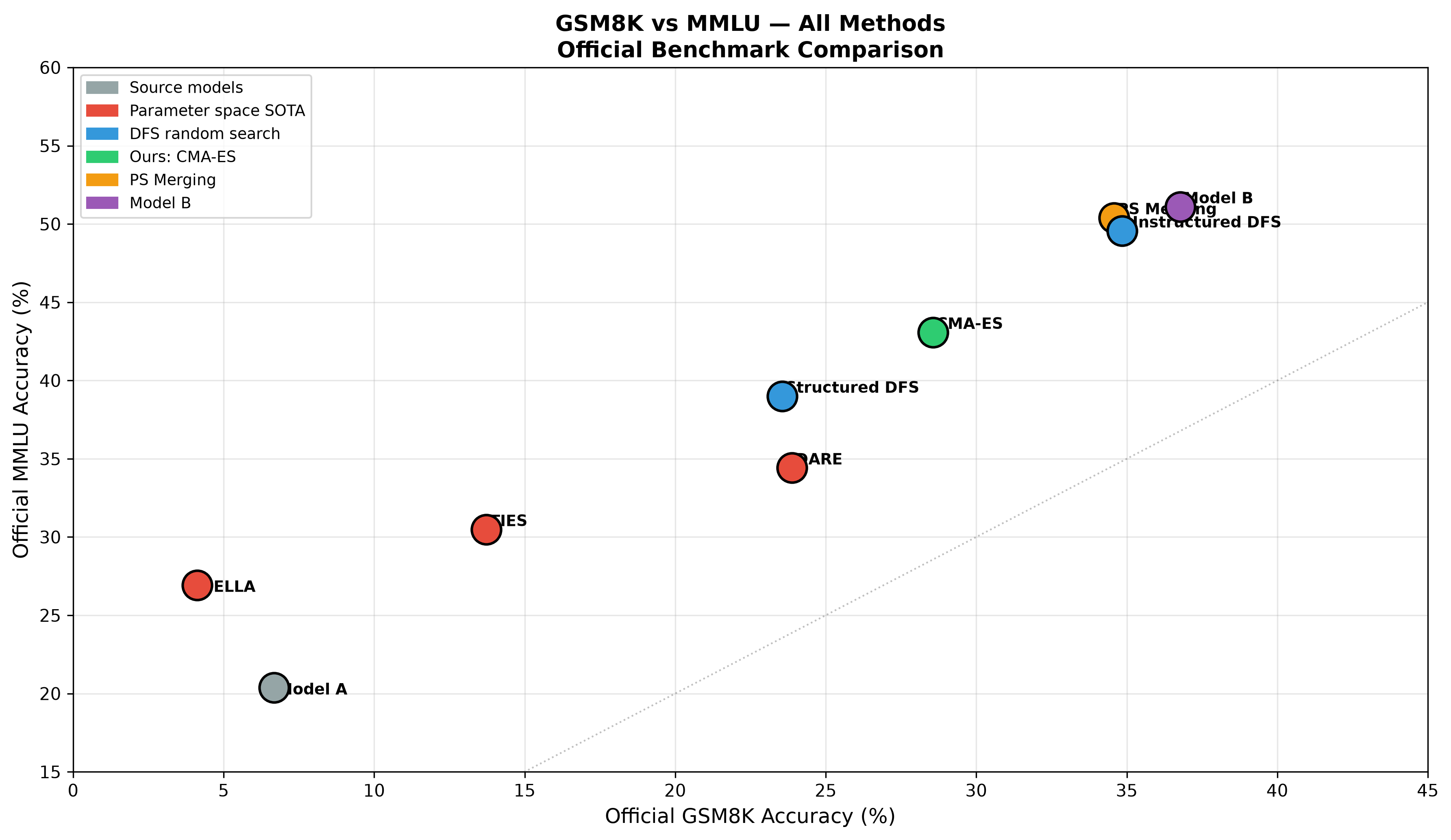}
    \caption{GSM8K vs MMLU scatter plot for all methods. Methods performing similarly on both benchmarks cluster near the diagonal. PS Merging and Unstructured DFS maintain competitive performance on both benchmarks, while CMA-ES shows stronger relative performance on MMLU than GSM8K, reflecting proxy optimization on GSM8K questions.}
    \label{fig:cross_benchmark}
\end{figure}

Figure~\ref{fig:cross_benchmark} reveals an important pattern: methods optimized or designed for general parameter averaging (PS Merging, Unstructured DFS) maintain performance across both benchmarks, while methods involving complex manipulation (TIES, DELLA) fail on both. Our CMA-ES approach shows relatively stronger performance on MMLU, reflecting its proxy optimization on GSM8K questions.

\subsection{Development Subset Results}

\begin{table}[!htbp]
\centering
\caption{Development Subset Results (97Q GSM8K, 100Q MMLU, 5 runs) — used for ablation only}
\label{tab:subset}
\begin{tabular}{lcc}
\toprule
Method & GSM8K Subset & MMLU Subset \\
\midrule
Model A & 43.8\% & 89.0\% \\
Model B & 85.4\% & 86.0\% \\
PS Merging & 84.4\% $\pm$ 0.0\% & 93.0\% $\pm$ 0.0\% \\
Unstructured DFS & 86.0\% $\pm$ 0.5\% & 91.6\% $\pm$ 1.2\% \\
Structured DFS (Random) & 86.5\% $\pm$ 1.1\% & --- \\
TIES~\cite{yadav2023} & 53.1\% $\pm$ 0.0\% & 90.0\% $\pm$ 0.0\% \\
DARE~\cite{yu2023} & 77.3\% $\pm$ 5.2\% & 92.2\% $\pm$ 2.5\% \\
DELLA~\cite{deep2024} & 53.1\% $\pm$ 2.4\% & 69.2\% $\pm$ 4.8\% \\
\textbf{Structured DFS + CMA-ES (Ours)} & \textbf{88.1\% $\pm$ 1.1\%} & \textbf{93.2\% $\pm$ 1.2\%} \\
\bottomrule
\end{tabular}
\end{table}

\begin{figure}[!htbp]
    \centering
    \includegraphics[width=\linewidth]{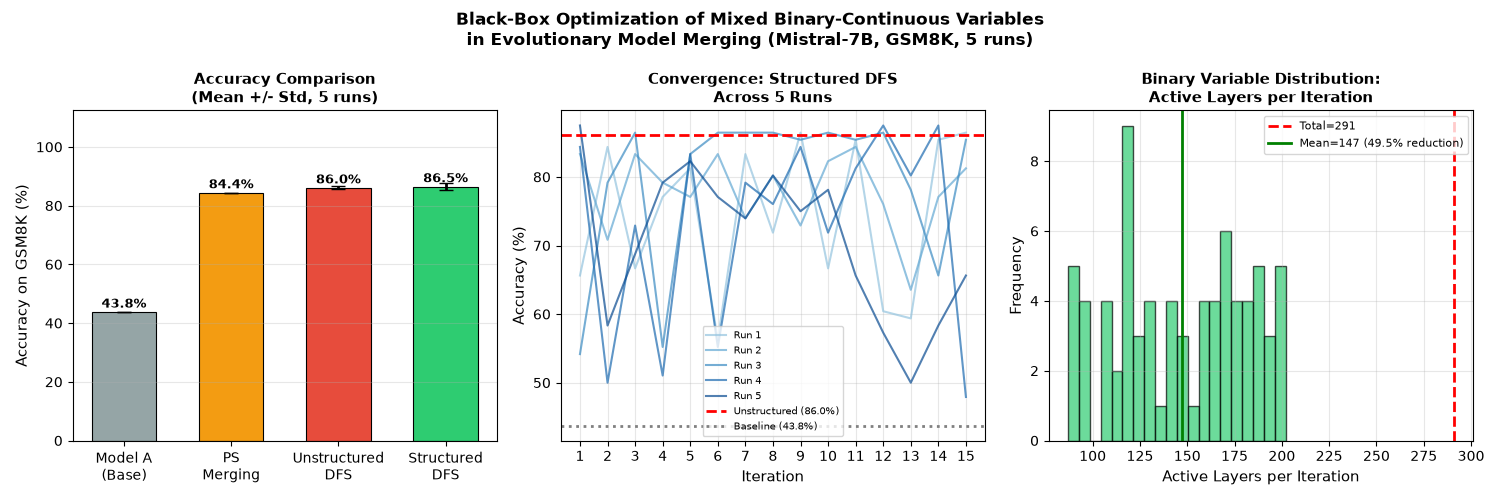}
    \caption{GSM8K development subset results (97 questions, 5 runs). Left: accuracy comparison. Center: CMA-ES convergence trajectories — volatile behavior confirms conditional dependency structure. Right: active layer distribution confirming 49.5\% search space reduction.}
    \label{fig:subset_results}
\end{figure}

\begin{figure}[!htbp]
    \centering
    \includegraphics[width=\linewidth]{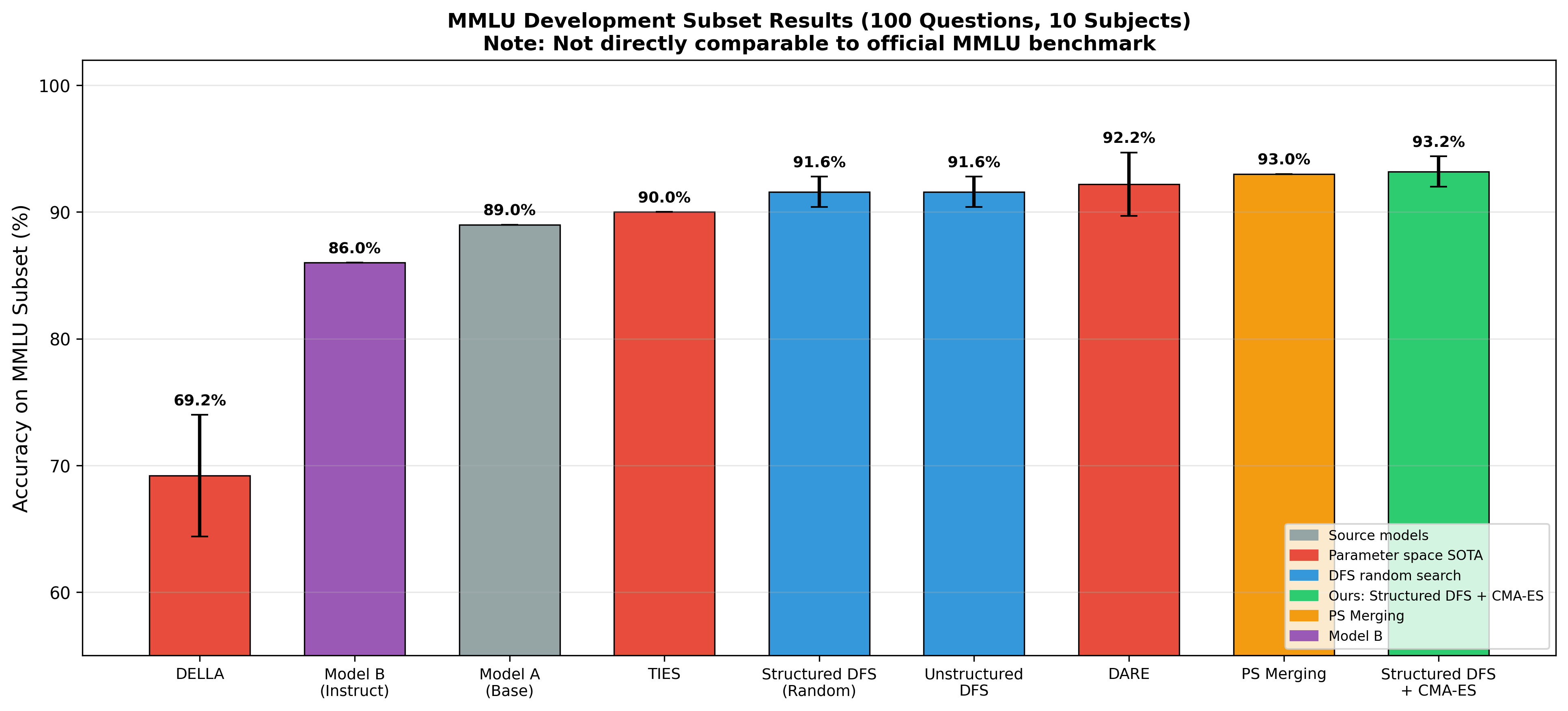}
    \caption{MMLU development subset results (100 questions, 10 subjects, 5 runs). CMA-ES achieves 93.2\% --- best across all methods on the development subset. Note: these results use a controlled subset and are not directly comparable to official benchmark results.}
    \label{fig:mmlu_subset}
\end{figure}

\subsection{Statistical Analysis}

With $n=3$ runs per method on official benchmarks, the Wilcoxon signed-rank test cannot achieve $p<0.05$ regardless of effect size (minimum achievable $p=0.125$). We therefore report Cohen's $d$ as the primary measure of practical significance.

\begin{table}[!htbp]
\centering
\caption{Effect Sizes (Cohen's $d$) — CMA-ES vs Other Methods}
\label{tab:stats}
\begin{tabular}{lccccc}
\toprule
& \multicolumn{2}{c}{Official GSM8K} & \multicolumn{2}{c}{Official MMLU} \\
\cmidrule(r){2-3} \cmidrule(r){4-5}
Comparison & Diff & Cohen's $d$ & Diff & Cohen's $d$ \\
\midrule
CMA-ES vs DELLA & +24.4\% & 13.05 & +16.1\% & 8.04 \\
CMA-ES vs TIES & +14.8\% & 7.97 & +12.6\% & 6.29 \\
CMA-ES vs DARE & +4.7\% & 2.08 & +8.7\% & 2.42 \\
CMA-ES vs Structured DFS & +5.0\% & 1.19 & +4.1\% & 1.12 \\
CMA-ES vs PS Merging & -6.0\% & -3.23 & -7.3\% & -3.66 \\
CMA-ES vs Unstr. DFS & -6.3\% & -3.08 & -6.5\% & -3.09 \\
\bottomrule
\end{tabular}
\end{table}

CMA-ES substantially outperforms TIES and DELLA (Cohen's $d > 6$ on both benchmarks), confirming that our method provides practically meaningful improvements over failing SOTA approaches. CMA-ES is outperformed by PS Merging and Unstructured DFS (Cohen's $d \approx -3$).

\subsection{Search Space Analysis}

\begin{figure}[!htbp]
    \centering
    \includegraphics[width=\linewidth]{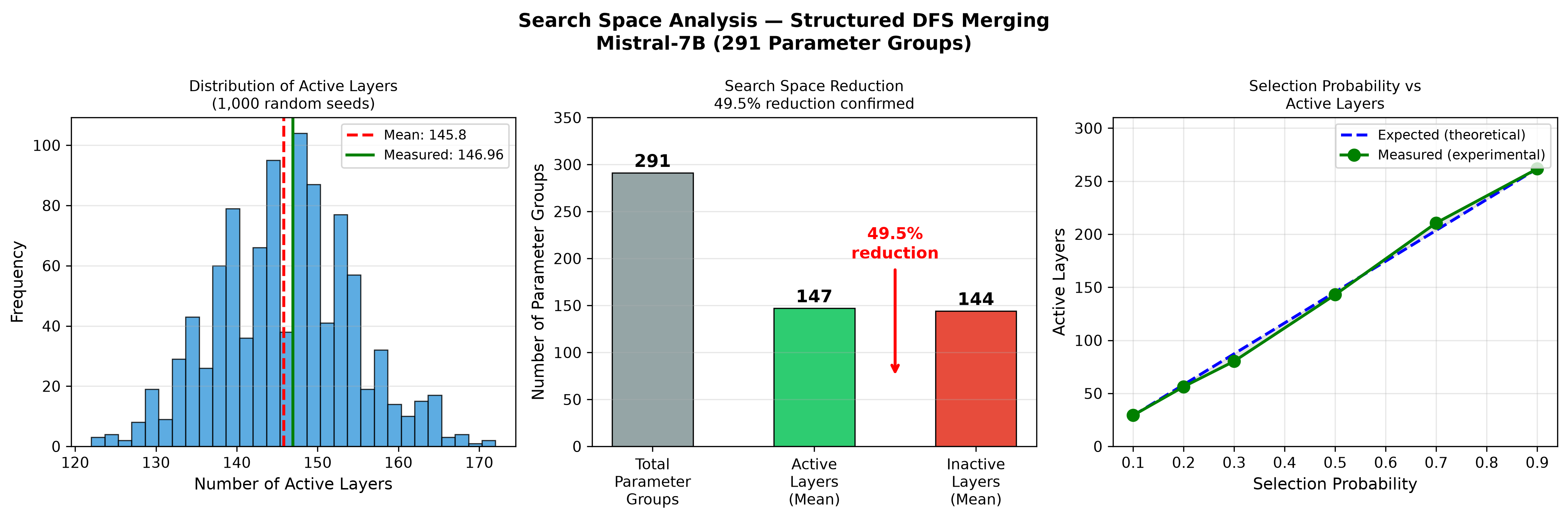}
    \caption{Search space analysis. Left: distribution of active layers across 1,000 random seeds (mean 146.96 of 291). Center: search space reduction visualization — 49.5\% reduction confirmed. Right: selection probability vs active layers showing theoretical and measured agreement.}
    \label{fig:search_space}
\end{figure}

Our structured approach reduces the effective continuous search space from 291 to a mean of 146.96 active layers — a 49.5\% reduction. This directly validates our theoretical formulation of the conditional dependency structure.

\subsection{CMA-ES Convergence}

\begin{figure}[!htbp]
    \centering
    \includegraphics[width=\linewidth]{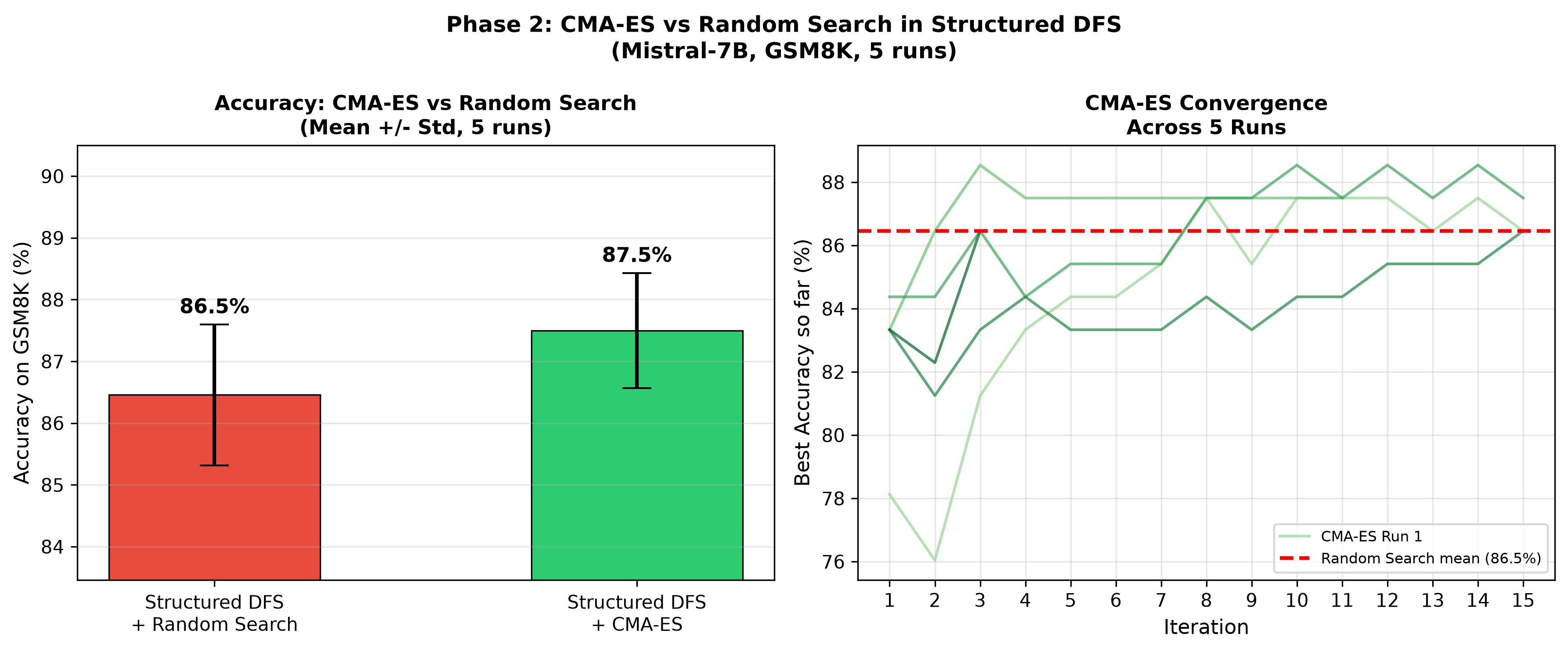}
    \caption{CMA-ES convergence on GSM8K development subset (97 questions, 5 runs). Left: accuracy comparison — CMA-ES (88.1\%) vs random search (86.5\%). Right: convergence trajectories showing systematic improvement. Red dashed line indicates random search mean.}
    \label{fig:convergence}
\end{figure}

\subsection{Per-Category MMLU Analysis}

\begin{figure}[!htbp]
    \centering
    \includegraphics[width=\linewidth]{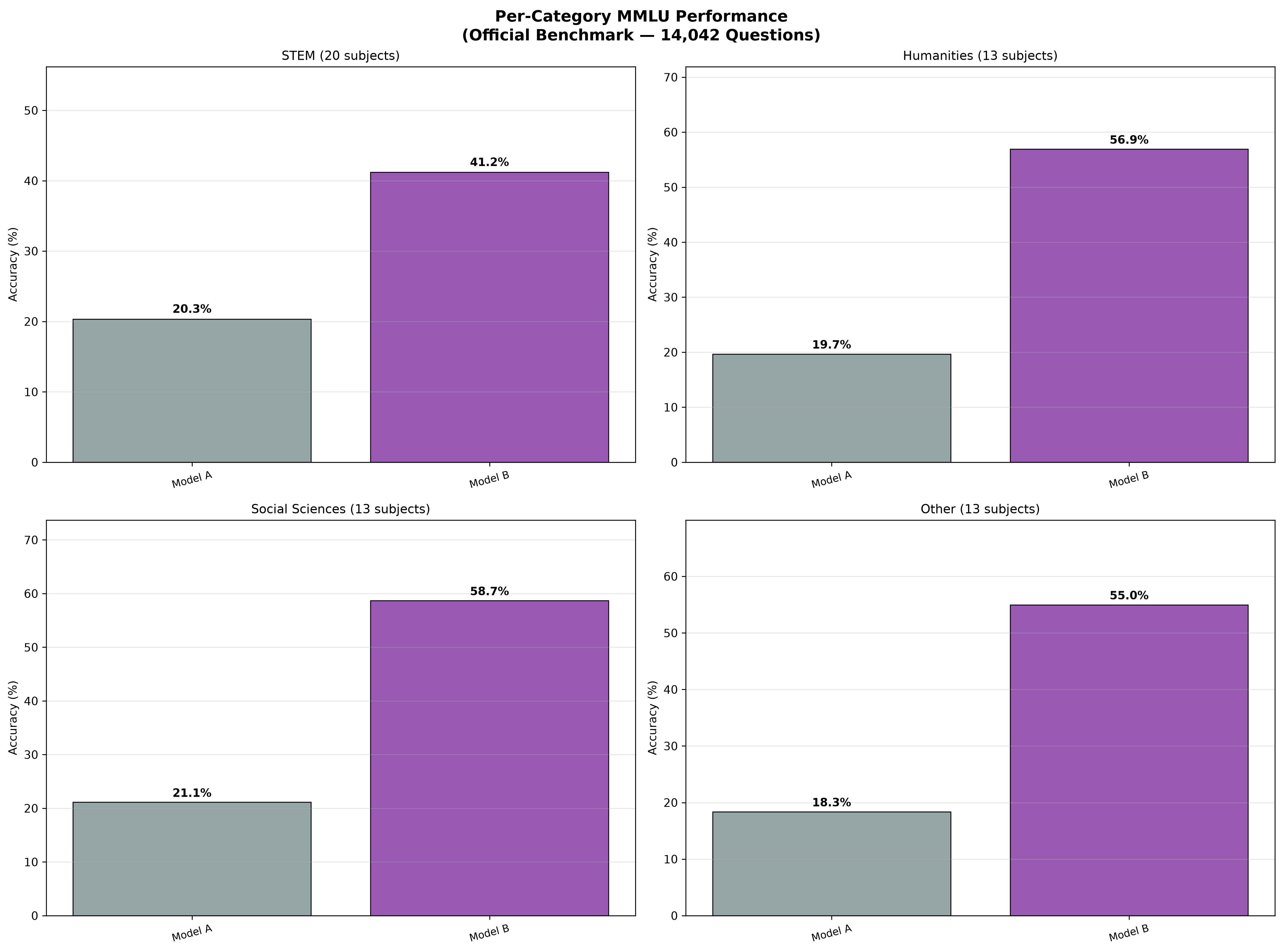}
    \caption{Per-category MMLU performance for Model A and Model B across STEM, Humanities, Social Sciences, and Other categories. Model B shows strongest performance on Social Sciences (58.7\%) and weakest on STEM (41.2\%), suggesting instruction tuning disproportionately benefits social and humanities reasoning.}
    \label{fig:per_category}
\end{figure}

\section{Ablation Study}

All ablation experiments conducted on development subsets. Full benchmark ablation is ongoing and will be reported in future work.

\subsection{Effect of Binary Selection Probability}

\begin{table}[!htbp]
\centering
\caption{Effect of Layer Selection Probability (GSM8K subset, 3 runs)}
\label{tab:abl1}
\begin{tabular}{lccc}
\toprule
Selection Prob & Active Layers & Mean Accuracy & Std \\
\midrule
0.1 & 29/291 (10\%) & 51.7\% & 6.9\% \\
0.2 & 56/291 (19\%) & 64.6\% & 9.0\% \\
0.3 & 80/291 (28\%) & 70.8\% & 7.3\% \\
0.5 & 143/291 (49\%) & 83.0\% & 3.5\% \\
0.7 & 211/291 (73\%) & 85.8\% & 0.5\% \\
0.9 & 262/291 (90\%) & 86.8\% & 0.5\% \\
\bottomrule
\end{tabular}
\end{table}

Variance decreases monotonically with selection probability — directly demonstrating Type-I interaction: sparse binary selections create stronger conditional dependencies and higher performance variance.

\subsection{Effect of Alpha Range}

\begin{table}[!htbp]
\centering
\caption{Effect of Continuous Weight Range (GSM8K subset, 3 runs)}
\label{tab:abl2}
\begin{tabular}{lcc}
\toprule
Alpha Range & Mean Accuracy & Std \\
\midrule
$[0.1, 0.9]$ & 86.1\% & 1.0\% \\
$[0.2, 0.8]$ & 83.3\% & 3.9\% \\
$[0.3, 0.7]$ & 82.6\% & 4.0\% \\
$[0.4, 0.6]$ & 79.5\% & 3.2\% \\
$[0.5, 0.5]$ & 74.0\% & 5.6\% \\
\bottomrule
\end{tabular}
\end{table}

\subsection{Fixed Alpha vs CMA-ES Learned Alpha}

\begin{table}[!htbp]
\centering
\caption{Fixed Alpha vs CMA-ES Learned Alpha (GSM8K + MMLU subsets, 3 runs)}
\label{tab:abl3}
\begin{tabular}{lcccc}
\toprule
& \multicolumn{2}{c}{GSM8K Subset} & \multicolumn{2}{c}{MMLU Subset} \\
\cmidrule(r){2-3} \cmidrule(r){4-5}
Method & Mean & Std & Mean & Std \\
\midrule
Fixed $\alpha = 0.7$ & 59.7\% & 6.0\% & 36.3\% & 6.8\% \\
Fixed $\alpha = 0.5$ & 74.0\% & 5.6\% & 67.7\% & 18.3\% \\
Fixed $\alpha = 0.3$ & 84.7\% & 1.3\% & 82.3\% & 11.9\% \\
\textbf{CMA-ES Learned} & \textbf{86.1\%} & \textbf{1.0\%} & 74.0\% & 13.6\% \\
\bottomrule
\end{tabular}
\end{table}

\begin{figure}[!htbp]
    \centering
    \includegraphics[width=\linewidth]{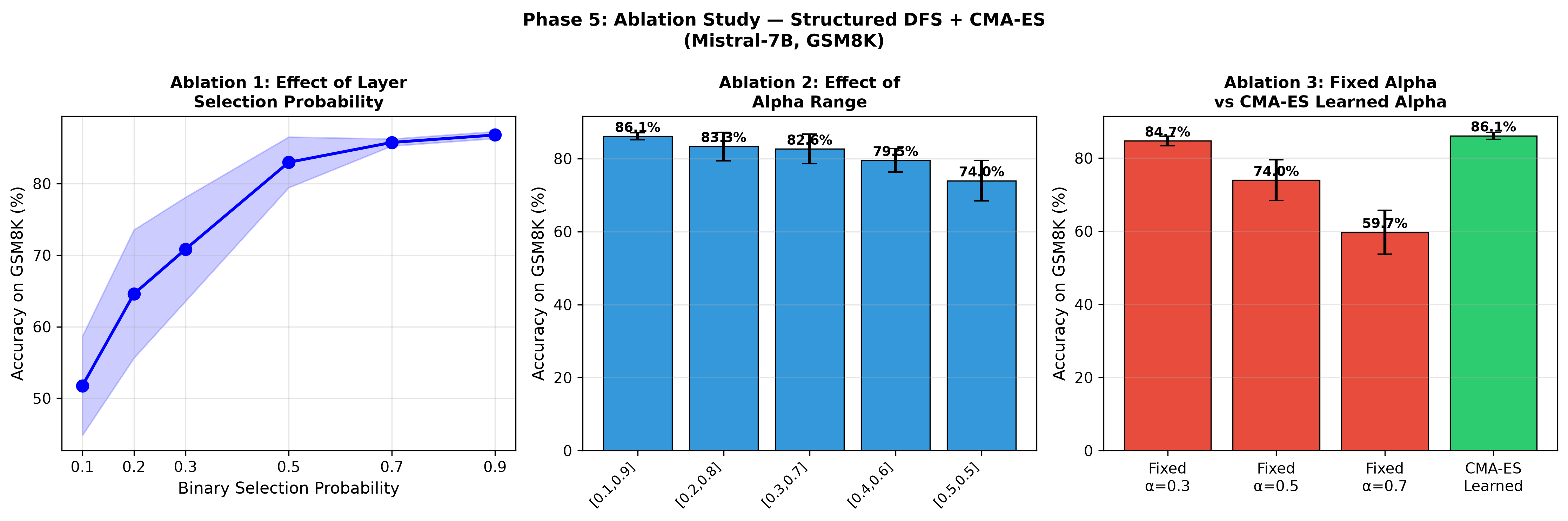}
    \caption{Ablation study results on GSM8K development subset. Left: selection probability effect. Center: alpha range effect. Right: fixed vs learned alpha — CMA-ES (86.1\% $\pm$ 1.0\%) outperforms all fixed values with lowest variance.}
    \label{fig:ablation}
\end{figure}

CMA-ES learned alpha outperforms all fixed values on GSM8K subset and achieves the lowest variance. On MMLU subset, fixed $\alpha=0.3$ (82.3\%) outperforms CMA-ES (74.0\%) — revealing a \textbf{proxy-task dependency}: CMA-ES optimized on GSM8K questions does not generalize to MMLU. This finding motivates multi-objective optimization as future work.

\section{Discussion}

\subsection{Why All Methods Fail on Official Benchmarks}

Our most important finding is that no merging method consistently surpasses Model B on either official benchmark. We identify three contributing factors:

\textbf{Factor 1 — Weak base model:} Mistral-7B-v0.1 scores only 6.67\% on GSM8K — far below Model B (36.77\%). The base model contributes little positive signal to the merged model for math reasoning.

\textbf{Factor 2 — Large weight distribution gap:} The parameter distributions of a base model and an instruction-tuned model differ substantially. Methods designed for merging same-family fine-tuned models (TIES, DELLA) fail catastrophically because their trimming and dropping operations destroy essential instruction-following weights.

\textbf{Factor 3 — Proxy evaluation mismatch:} CMA-ES optimization on a 100-question proxy does not fully capture the requirements of 1,319 GSM8K or 14,042 MMLU questions. This explains why CMA-ES performs well on the development subset but not on official benchmarks.

\subsection{Why TIES and DELLA Fail So Dramatically}

TIES (13.72\% on GSM8K) and DELLA (4.12\% on GSM8K) perform catastrophically. These methods were designed for merging fine-tuned models sharing the same base. When merging a base model with an instruction-tuned model, the delta weights are large and fundamental — trimming or dropping them destroys essential knowledge.

\subsection{Type-I Interaction Validation}

Our ablation results directly validate the Type-I interaction structure. In Ablation 1, performance variance decreases monotonically with selection probability — exactly what Type-I interaction predicts: when fewer continuous dimensions are active, small changes in binary selection create large performance differences.

\subsection{Connection to Concurrent Work}

Ong et al.~\cite{ong2026igbd} independently developed IGBD — a bilevel optimization framework designed specifically for Type-I interaction. Their work validates our theoretical analysis: the interaction structure we identify in DFS merging is a genuine algorithmic challenge requiring specialized methods. Applying IGBD to DFS merging represents the natural next step, and we expect it to substantially improve over our CMA-ES baseline.

\subsection{Limitations}

\begin{enumerate}
    \item \textbf{Single model family:} Experiments limited to Mistral-7B. Results may differ for other architectures.
    \item \textbf{Weak base model:} Mistral-7B-v0.1 at 6.67\% GSM8K limits merging potential. Stronger base models may yield different results.
    \item \textbf{Simplified CMA-ES:} Binary selection is random, not jointly optimized. IGBD~\cite{ong2026igbd} represents the theoretically correct approach.
    \item \textbf{Proxy evaluation gap:} CMA-ES optimized on small proxy subset does not generalize well to full benchmarks.
    \item \textbf{Statistical power:} 3 runs per method limits statistical testing.
\end{enumerate}

\section{Conclusion}

We formally characterize Data Flow Space merging as a mixed binary-continuous black-box optimization problem exhibiting Type-I interaction. Our comprehensive evaluation on official GSM8K and MMLU benchmarks reveals that all tested merging methods fail to surpass the stronger source model — with TIES and DELLA performing catastrophically, and even DELLA falling below the base model on GSM8K. These results directly validate the fundamental difficulty of the optimization structure we identify.

Our findings point clearly toward IGBD~\cite{ong2026igbd} as the theoretically appropriate algorithm for this problem, multi-objective merging formulations to balance task-specific and general knowledge, and the importance of stronger base model selection.

\section*{Acknowledgments}
The author thanks Professor Youhei Akimoto for valuable discussions and feedback on earlier versions of this research that shaped the direction of this work, and for his foundational contributions to mixed categorical-continuous optimization theory.

\section*{Data availability}
All experimental results and code are publicly available at \url{https://github.com/Mdniloykhan/dfs-merging-blackbox-optimization} and \url{https://zenodo.org/records/20123507}.

\section*{Conflict of interest}
The author declares no conflict of interest.

\end{document}